# Parsimonious Random Vector Functional Link Network for Data Streams

Mahardhika Pratama, Plamen P. Angelov, Edwin Lughofer

*Abstract* – the theory of random vector functional link network (RVFLN) has provided a breakthrough in the design of neural networks (NNs) since it conveys solid theoretical justification of randomized learning. Existing works in RVFLN are hardly scalable for data stream analytics because they are inherent to the issue of complexity as a result of the absence of structural learning scenarios. A novel class of RVLFN, namely parsimonious random vector functional link network (pRVFLN), is proposed in this paper. pRVFLN adopts a fully flexible and adaptive working principle where its network structure can be configured from scratch and automatically generated in accordance with nonlinearity and time-varying property of system being modelled. pRVFLN is equipped with complexity reduction scenarios where inconsequential hidden nodes can be pruned and input features can be dynamically selected. pRVFLN puts into perspective an online active learning mechanism which expedites the training process and relieves operator's labelling efforts. In addition, pRVFLN introduces a non-parametric type of hidden node, developed using an interval-valued data cloud. The hidden node completely reflects the real data distribution and is not constrained by a specific shape of the cluster. All learning procedures of pRVFLN follow a strictly single-pass learning mode, which is applicable for online time-critical applications. The advantage of pRVFLN was verified through numerous simulations with real-world data streams. It was benchmarked with recently published algorithms where it demonstrated comparable and even higher predictive accuracies while imposing the lowest complexities. Furthermore, the robustness of pRVFLN was investigated and a new conclusion is made to the scope of random parameters where it plays vital role to the success of randomized learning.

*Keywords – Random Vector Functional Link, Evolving Intelligent System, Online Learning, Randomized Neural Networks*

## I. Introduction

For decades, research in artificial neural networks has mainly investigated the best way to determine network free parameters, which reduces error as close as possible to zero [18]. Various approaches [55], [56] were proposed, but a large volume of work is based on a first or second-order derivative approach in respect to the loss function [19], [36]. Due to the rapid technological progress in data storage, capture, and transmission [34], the machine learning community has encountered an information explosion, which calls for scalable data analytics. Significant growth of the problem space has led to a scalability issue for conventional machine learning approaches, which require iterating entire batches of data over multiple epochs. This phenomenon results in a strong demand for a simple, fast machine learning algorithm to be well-suited for deployment in numerous data-rich applications [58]. This provides a strong case for research in the area of randomized neural networks (RNNs) [10], [13], [52], which was very popular in late 80's and early 90's. RNNs offer an algorithmic framework, which allows them to generate most of the network parameters randomly while still retaining reasonable performance [8], [47]. One of the most prominent RNNs in the literature is the random vector functional link (RVFL) network which features solid universal approximation theory under strict conditions [10].

Due to its simple but sound working principle, the RNN has regained its popularity in the current literature [8], [11], [14], [15], [17], [41], [50], [51], [63], [66], [67]. Nonetheless, vast majority of RNNs in the literature suffer from the issue of complexity which make their computational complexity and memory burden prohibitive for data stream analytics since their complexities are manually determined and rely heavily on expert domain knowledge [12]. They presents a fixed-size structure which lacks of adaptive mechanism to encounter changing training patterns in the data streams [15]. Random selection of network parameters often leads the network complexity to go beyond what necessary due to existence of superfluous hidden nodes which contribute little to the generalization performance [45]. Although the universal approximation capability of RNN is assured only when a sufficient complexity is selected, choosing suitable complexity for given problems entails expert-domain knowledge and is problem-dependent.

A novel RVFLN, namely parsimonious random vector functional link network (pRVFLN), is proposed. pRVFLN combines simple and fast working principles of RFVLN where all network parameters but the output weights are randomly generated with no tuning mechanism of hidden nodes. It characterises the online and adaptive nature of evolving intelligent systems where network components are automatically generated on the fly. pRVFLN is capable of following any variations of data streams no matter how slow, rapid, gradual, sudden or temporal drifts in data streams are because it can initiate its learning structure from scratch with no initial structure and its structure is self-evolved from data streams in the one-pass learning mode by automatically adding, pruning and recall its hidden nodes [45]. Furthermore, it is compatible for online real-time deployment because data streams are handled without revising previously seen samples. pRVFLN is equipped with the hidden node pruning mechanism which guarantees low structural burdens and the rule recall mechanism which aims to address the cyclic concept drift. pRVFLN incorporates a dynamic input selection scenario which makes possible activation and deactivation of input attributes on the fly and an online active learning scenario which rules out inconsequential samples from training process. Moreover, pRVFLN is a plug-and-play scenario where a single training process encompasses all learning scenarios and a sample-wise manner without pre-and/or post-processing steps. Novelties of pRVFLN are elaborated as follows:

- *Network Architecture*: Unlike majority of existing RVFLNs, pRVFLN is structured as a locally recurrent neural network with a self-feedback loop at the hidden node to capture temporal system dynamic [24]. The recurrent network architecture puts into perspective a spatio-temporal activation degree which takes into account compatibility of both previous and current data points without compromising the local learning property because of its local recurrent connection. pRVFLN introduces a new type of hidden node, namely an interval-valued data cloud inspired by the notion of recursive density estimation and AnYa by Angelov [1], [2]. This hidden node is parameter-free and requires no parameterization per scalar variable. It is not constrained by a specific shape and its firing strength is defined as a local density calculated as accumulated distances between a local mean and all data points in the cloud seen thus far. Our version in this paper distinguishes itself from its predecessors in [1], [2] in the fact that an interval uncertainty is incorporated per local region which targets imprecision and uncertainties of sensory data [50]. Interval-valued data cloud still satisfies the universal approximation condition of the RVFLN since it is derived using the Cauchy kernel which is asymptotically a Gaussian-like function. Moreover, the output layer consists of a collection of functional-link output nodes created by the second order Chebyshev polynomial which increases the degree of freedom of the output weight to improve its approximation power [38].

- *Online Active Learning Mechanism*: pRVFLN possesses an online active learning mechanism which is meant to extract data samples for training process. This mechanism is capable of finding important data streams for training process, while discarding inconsequential samples [54]. This strategy expedites the training mechanism and improves model's generalization since it prevents redundant samples to be learned. This scenario is underpinned by sequential entropy method (SEM) which forms a modified version of neighbourhood probability method [64] for data streams. The SEM quantifies the entropy of neighbourhood probability recursively and differs from that of [45] because it integrates the concept of data cloud. The concept of data cloud further simplifies the working principle of SEM since its activation degree already abstracts a density of a local region – a key attribute of neighbourhood probability. In realm of classification problems, the SEM does not always call for ground truth which leads to significant reduction of operator annotation's efforts.
- *Hidden Node Growing Mechanism*: pRVFLN is capable of automatically generating its hidden nodes on the fly with the help of type-2 self-constructing clustering (T2SCC) mechanism which was originally designed for an incremental feature clustering mechanism [26], [32]. This method relies on the correlation measure in the input space and target space, which can be carried out in the single-pass learning mode with ease. The original version, cannot be directly implemented here because the original version is not yet designed for the interval-valued data cloud. This rule growing process also differs from those of other RNNs with a dynamic structure [12], because the hidden nodes are not randomly generated, rather they are created from the rule growing condition, which considers the locations of data samples in the input space. We argue that randomly choosing centers or focal points of hidden nodes independently from the original training data risks on performance deterioration [57], because it may not hold the completeness principle. pRVFLN relies on the data cloud-based hidden node, which does not require any parameterization, thus offering a simple and fast working framework as other RNNs.

- *Hidden Node Pruning and Recall Mechanism*: a rule pruning scenario is integrated in pRVFLN. The rule pruning scenario plays vital role to assure modest network structures since it is capable of detecting superfluous neurons to be removed during the training process [34]. pRVFLN employs the so-called type-2 relative mutual information (T2RMI) method which extends the RMI method [21] to the working principle of the interval-valued hidden node. The T2RMI method examines relevance of hidden nodes to target concept and thus captures outdated hidden nodes which is no longer compatible to portray the current target concept. In addition, the T2RMI method is applied to recall previously pruned rules. This strategy is important to deal with the so-called recurring drift. The recurring drift refers to a situation where old concept re-emerges again in the future. The absence of rule recall scenario risks on the catastrophic forgetting of previously valid knowledge because the cyclic drift imposes an introduction of a new rule without any memorization of learning history. It differs from that [34] because the rule recall mechanism is separated from the rule growing scenario.
- *Online Feature Selection Mechanism*: pRVFLN is capable of carrying out an online feature selection process, borrowing several concepts of online feature selection (OFS) [59]. Note that although feature selection is well-established for an offline situation, online feature selection remains a challenging and unsolved problem because feature contribution must be measured with the absence of a complete dataset. Notwithstanding some online feature reduction scenarios in the literature, the issue of stability is still open, because an input feature is permanently forgotten once pruned. The OFS delivers a flexible online feature selection scenario, because it makes possible to select or deselect input attributes in every training observation by assigning crisp weights (0 or 1) to input features. Another prominent characteristic of the OFS method lies in its capability to deal with partial input features, because the cost of extracting all input features may be expensive in certain applications. The convergence of the OFS for both full and partial input attributes have been also proven. Nevertheless, the original version [59] is originally devised for linear regression and calls for some modification to be perfectly fit to pRVLFN. The generalized OFS (GOFS) is proposed here and incorporates some refinements to adapt to the working principle of the pRVFLN.

This paper puts forward the following contributions: 1) it presents a novel random vector functional link network termed pRVFLN. pRVFLN aims to handle the issue of data streams which remains an uncharted territory of current RVFLN works; 2) it puts into perspective the interval-valued data cloud which is not shape-specific and non-parametric. It requires no parameterization per scalar variable and makes use of local density information of a data cloud. This strategy aims to hinder complete randomization of hidden node parameters while still maintaining no tuning principle of RVFLN. Randomization of all hidden node parameters blindly without bearing in mind true data distribution lead to common pitfall of RVFL where the hidden node matrix tend to be ill-defined and compromises the completeness of network structure as confirmed in [57]. ; 3) four learning components, namely SEM, T2SCC, T2RMI, GOFS, are proposed. The efficacy of pRVFLN was thoroughly evaluated using numerous real-world data streams and benchmark with recently published algorithms in the literature where pRVFLN demonstrated a highly scalable approach for data stream analytics with trivial cost of its generalization performance. A supplemental document containing additional numerical studies, analysis of predefined threshold and the effect of learning components is also provided in[1]. Moreover, analysis of robustness of random intervals was performed. It is concluded that random regions should be carefully selected and should be chosen close enough to true operating regions of a system being modelled. The MATLAB codes of pRVFLN is made publicly available in [4] to help further study.

The rest of this paper is structured as follows: Network architecture of pRVFLN is outlined in Section 2; Algorithmic development of pRVFLN is detailed in Section 3; Proof of concepts are outlined in Section 4; Conclusions are drawn in the last section of this paper.

## II. Network Architecture of pRVFLN

pRVFLN utilises a local recurrent connection at the hidden node which generates the spatiotemporal property. In the literature, there exist at least three types of recurrent network structures referring to its recurrent connections: global [22], [24], interactive [33], and local [23], but the local recurrent connection is deemed to be the most compatible recurrent type to our case because it does not harm the local property, which assures stability when adding, pruning and fine-tuning hidden nodes. pRVFLN utilises the notion of the functional-link neural network which presents the so-called enhancement layer. Furthermore, the hidden layer of pRVFLN is built upon an interval-valued data cloud, which does not require any parametrization and granulation [1]. This hidden node does not have any pre-specified shape and fully evolves its shape with respect to the real data distribution because it is derived from recursive local density estimation. We bring the idea of data cloud further by integrating an interval-valued principle.

Suppose that a data tuple streams at *t-the* time instant, where $X_t \in \Re^n$ is an input vector and $T_t \in \Re^m$ is a target vector while $n$ and $m$ are respectively the numbers of input and output variables. Because pRVFLN works in a strictly online learning environment, it has no access to previously seen samples, and a data point is simply discarded after being learned. Due to the pre-requisite of an online learner, the total number of data $N$ is assumed to be unknown. The output of pRVFLN is defined as follows:

$$y_o = \sum_{i=1}^{R} \beta_i \tilde{G}_{i,temporal}(A_t X_t + B_t), \tilde{G}_{temporal} = [\underline{G}, \overline{G}] \tag{1}$$

where $R$ denotes the number of hidden nodes. $\beta_i$ stands for the *i-th* output node, produced by weighting the weight vector with an extended input vector $\beta_i = x_e^T w_i$, where $x_e \in \Re^{(2n+1)\times 1}$ is an extended input vector resulting from the functional link neural network based on the Chebyshev function and $w_i \in \Re^{(2n+1)\times 1}$ is a connective weight of the *i-th* output node. $A_t \in \Re^n$ is an input weight vector randomly generated here. $B_t$ is the bias fixed at 0 for simplicity. $\tilde{G}_{i,temporal}$ is the *i-th* interval-valued data cloud, triggered by the upper and lower data cloud $\underline{G}_{i,temporal}, \overline{G}_{i,temporal}$. By expanding the interval-valued data cloud, the following is obtained:

$$y_o = \sum_{i=1}^{R}(1-q_o)\beta_i \overline{G}_{i,temporal} + \sum_{i=1}^{R} q_o \beta_i \underline{G}_{i,temporal} \tag{2}$$

where $q \in \Re^m$ is a design factor to reduce an interval-valued function to a crisp one [9]. It is worth noting that the upper and lower activation functions $\underline{G}_{i,temporal}, \overline{G}_{i,temporal}$ deliver spatiotemporal characteristics as a result of a local recurrent connection at the *i-th* hidden node, which combines the spatial and temporal firing strength of the *i-th* hidden node. These temporal activation functions output the following.

$$\underline{G}_{i,temporal}^{t} = \lambda_i \underline{G}_{i,spatial}^{t} + (1-\lambda_i)\underline{G}_{i,temporal}^{t-1}, \overline{G}_{i,temporal}^{t} = \lambda_i \overline{G}_{i,spatial}^{t} + (1-\lambda_i)\overline{G}_{i,temporal}^{t-1}$$

where $\lambda \in \Re^R$ is a weight vector of the recurrent link. The local feedback connection here feeds the spatiotemporal firing strength at the previous time step $\tilde{G}_{i,temporal}^{t-1}$ back to itself and is consistent with the local learning principle. This trait happens to be very useful in coping with the temporal system dynamic because it functions as an internal memory component which memorizes a previously generated spatiotemporal activation function at *t-1*. Also, the recurrent network is capable of overcoming over-dependency on time-delayed input features and lessens strong temporal dependencies of subsequent patterns [36]. This trait is desired in practise since it may lower input dimension, because prediction is done based on recent measurement only. Conversely, the feedforward network assumes a problem as a function of current and past input and outputs. This strategy at least entails expert knowledge for system order to determine the number of delayed components.

The hidden node of the pRVFLN is an extension of the cloud-based hidden node, where it embeds an interval-valued concept to address the problem of uncertainty [30]. Instead of computing an activation degree of a hidden node to a sample, the cloud-based hidden node enumerates the activation degree of a sample to all intervals in a local region on-the-fly. This results in local density information, which fully reflects real data distributions. This concept was defined in AnYa [1], [3] and was patented in [2]. This concept is also the underlying component of AutoClass and TEDA-Class [4]. All of which come from Angelov's sound work of RDE [3]. This paper aims to modify these prominent works to the interval-valued case. Suppose that $N_i$ denotes the support of the *i-th* data cloud, an activation degree of *i-th* cloud-based hidden node refers to its local density estimated recursively using the Cauchy function:

$$\tilde{G}_{i,spatial} = \frac{1}{1+\sum_{k=1}^{N_i}\left(\frac{\tilde{x}_k - x_t}{N_i}\right)^2} \tag{3}$$

$\tilde{x}_{k,i} = [\underline{x}_{k,i}, \overline{x}_{k,i}], \tilde{G}_{i,spatial} = [\underline{G}_{i,spatial}, \overline{G}_{i,spatial}]$

where $\tilde{x}_k$ is *k-th* interval in the *i-th* data cloud and $x_t$ is *t-th* data sample. It is observed that Eq. (3) requires the presence of all data points seen so far, which is impossible when dealing with data streams. Its recursive form is formalised in [3] and is generalized here to the interval-valued problem:

$$\overline{G}_{i,spatial} = \frac{1}{1+\left\|A_t^T x_t - \overline{\mu}_{i,N_i}\right\|^2 + \overline{\Sigma}_{i,N_i} - \left\|\overline{\mu}_{i,N_i}\right\|^2}, \underline{G}_{i,spatial} = \frac{1}{1+\left\|A_t^T x_t - \underline{\mu}_{i,N_i}\right\|^2 + \underline{\Sigma}_{i,N_i} - \left\|\underline{\mu}_{i,N_i}\right\|^2} \quad (4)$$

where $\underline{\mu}_i, \overline{\mu}_i$ signify the upper and lower local means of the *i-th* cloud:

$$\underline{\mu}_{i,N_i} = \left(\frac{N_i - 1}{N_i}\right)\underline{\mu}_{i,N_i-1} + \frac{x_{i,N_i} - \Delta_i}{\left\|N_i\right\|}, \underline{\mu}_{i,1} = x_{i,1} - \Delta_i, \overline{\mu}_{i,N_i} = \left(\frac{N_i - 1}{N_i}\right)\overline{\mu}_{i,N_i-1} + \frac{x_{i,k} + \Delta_i}{\left\|N_{i,k}\right\|}, \overline{\mu}_{i,1} = x_{i,1} + \Delta_i \quad (5)$$

where $\Delta_i$ is an uncertainty factor of the *i-th* cloud, which determines the degree of tolerance against uncertainty. The uncertainty factor creates an interval of the data cloud, which controls the degree of tolerance for uncertainty. It is worth noting that a data sample is considered as a population of the *i-th* cloud when resulting in the highest density. Moreover, $\overline{\Sigma}_{i,N_i}, \underline{\Sigma}_{i,N_i}$ are the upper and lower mean square lengths of the data vector in the *i-th* cloud as follows:

$$\underline{\Sigma}_{i,N_i} = \left(\frac{N_i - 1}{N_i}\right)\underline{\Sigma}_{i,N_i-1} + \frac{\left\|x_{i,N_i}\right\|^2 - \Delta_i}{N_i}, \underline{\Sigma}_{i,1} = \left\|x_{i,1}\right\|^2 - \Delta_i, \overline{\Sigma}_{i,N_i} = \left(\frac{N_{i,k} - 1}{N_{i,k}}\right)\overline{\Sigma}_{i,N_i-1} + \frac{\left\|x_{i,N_i}\right\|^2 + \Delta_i}{N_i}, \overline{\Sigma}_{i,1} = \left\|x_{i,1}\right\|^2 + \Delta_i \quad (6)$$

It can be observed from Eq. (3) that the cloud-based hidden node does not have any specific shape and evolves naturally according to its supports. Furthermore, it is parameter-free, where no parameters – centroid, width, etc. as encountered in the conventional hidden node need to be assigned. Although the concept of the cloud-based hidden node was generalized in TeDaClass [27] by introducing the eccentricity and typicality criteria, the interval-valued idea is uncharted in [27]. Note that the Cauchy function is asymptotically a Gaussian-like function, satisfying the integreable requirement of the RVFLN to be a universal approximator.

Unlike conventional RNNs, pRVFLN puts into perspective a nonlinear mapping of the input vector through the Chebyshev polynomial up to the second order. This concept realises an enhancement layer linking the input layer to the output layer - consistent with the original concept of the RVFLN. Note that recently developed RVFLNs in the literature mostly neglect the direct connection because they are designed with a zero-order output node [8], [11], [14], [15], [17], [41], [50], [51], [63], [66],[67] . The direct connection expands the output node to a higher degree of freedom, which aims to improve the local mapping aptitude of the output node. The direct connection produces the extended input vector $x_e$ making use of the second order Chebyshev polynomial:

$$v_{p+1}(x) = 2x_j v_p(x_j) - v_{p-1}(x_j) \quad (7)$$

where $v_0(x_j) = 1, v_1(x_j) = x_j, v_2(x_j) = 2x_j^2 - 1$. Suppose that three input attributes are given $X = [x_1, x_2, x_3]$ , the extended input vector is expressed as the Chebyshev polynomial up to the second order $x_e = [1, x_1, \nu_2(x_1), x_2, \nu_2(x_2), x_3, \nu(x_3)]$. Note that the term 1 here represents an intercept of the output node. This avoids an output node from going through origin, which may risk an untypical gradient. There exist other functions as well in functional-link neural networks: trigonometric and polynomial, power. The Chebyshev function, however, scatters fewer parameters to be stored into memory than the trigonometric function, while the Chebyshev function has better mapping capability than other polynomial functions of the same order. In addition, the polynomial power function is not robust against an extrapolation case. pRVFLN implements the random learning concept of the RVFLN, in which all parameters, namely the input weight $A$, design factor $q$, recurrent link weight $\lambda$, and uncertainty factor $\Delta$, are randomly generated. Only the weight vector is left for parameter learning scenario $w_i$. Since the hidden node is parameter-free, no randomization takes place for hidden node parameters. The network structure of pRVFLN and the interval-valued data cloud are depicted in Fig. 1 and 2 respectively.

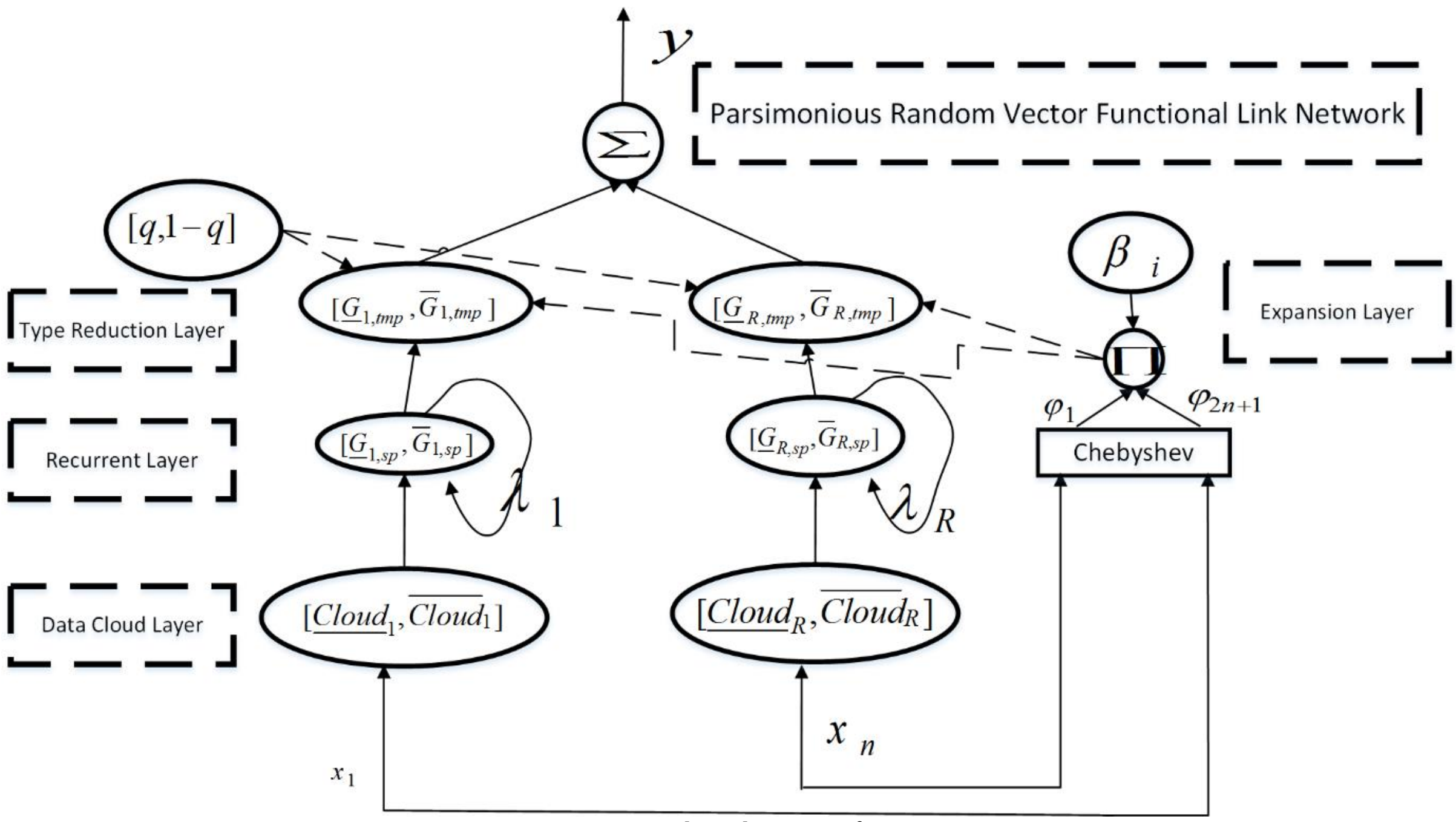

Fig. 1 Network Architecture of pRVFLN

## III. Learning Policy of pRVFLN

This section discusses the learning policy of pRVFLN. Section 3.1 outlines the online active learning strategy, which deletes inconsequential samples. Samples, selected in the sample selection mechanism, are fed into the learning process of pRVFLN. Section 3.2 deliberates the hidden node growing strategy of pRVFLN. Section 3.3 elaborates the hidden node pruning and recall strategy, while Section 3.4 concerns the online feature selection mechanism. Section 3.5 explains the parameter learning scenario of pRVFLN. Algorithm 1 shows the pRVFLN learning procedure.

### *3.1 Online Active Learning Strategy*

The active learning component of the pRVFLN is built on the extended sequential entropy (ESEM) method, which is derived from the SEM method [64]. The ESEM method makes use of the entropy of the neighborhood probability to estimate the sample contribution. There exist at least three salient facets that distinguish the ESEM from its predecessor [64]:1) it forms an online version of the SEM; 2) it is combined with the concept of the data cloud, which accurately represents the local density; and 3) it handles regression as well as classification because the sample contribution is enumerated without the presence of true class label. One may agree that the vast majority of sample selection variants are designed for classification problems only, and delve sample's location in respect to the decision surface. To the best of our knowledge, only Das et al. [16] address the regression problem, but it still shares the same principle as its predecessors, exploiting the hinge cost function to evaluate sample contribution [54]. The concept of neighborhood probability refers to the probability of an incoming data stream sitting in the existing data clouds, which is written as follows:

$$P(X_t \in N_i) = \frac{\sum_{k=1}^{N_i} \frac{M(X_t, x_k)}{N_i}}{\sum_{i=1}^{R} \sum_{k=1}^{N_i} \frac{M(X_t, x_k)}{N_i}} \tag{8}$$

Algorithm 1. Learning Architecture of pRVFLN

**Algorithm 1: Training Procedure of pRVFLN**

***Define:*** *Training Data* $(X_t, T_t) = (x_1,..,x_n,t_1,..,t_m)$
*Predefined Parameters* $\alpha_1, \alpha_2$
***/*Step 1: Online Active Learning Strategy/****
***For i=1 to R do***
  *Calculate the neighborhood probability (8) with spatial firing strength (4)*
***End For***
  *Calculate the entropy of neighborhood probability (8) and the ESEM (10)*
***IF (34) Then***
***/*Step 2: Online Feature Selection/****
***IF*** *Partial=Yes* ***Then***
  *Execute Algorithm 3*
***Else IF***
  *Execute Algorithm 2*
***End IF***
***/*Step 3: Data Cloud Growing Mechanism/****
***For j=1 to n do***
  *Compute* $\xi(x_j, T_o)$
***End For***
***For i=1 to R do***
  *Calculate input coherence (12)*
***For o=1 to m do***
  *Calculate* $\xi(\tilde{\mu}_i, T_o)$
***End For***
  *Compute the output coherence (13)*
***End For***
***IF (16) Then***
  *Associate a sample to a data cloud to the i* data cloud Ni*=Ni*+1*
  *Update the local mean and the square length of i*-th data cloud (5), (6)*
***Else***
  *Create new data cloud*
  *Take the next sample and Go to Phase 1*
***End IF***
***/*Step 4: Data Cloud Pruning Mechanism/****
***For i=1 to R do***
  ***For o=1 to m do***
    *Calculate* $\xi(\tilde{G}_{i,temp}, T_o)$
  ***End For***
  ***IF (19) Then***
   *Discard i-th data cloud*
  ***End IF***
***End For***
***/*Step 5: Adaptation of Output Weight/****
***For i=1 to R do***
  *Update output weights using FWGRLS*
***End For***

where $X_T$ is a newly arriving data point and $x_n$ is a data sample, associated with the *i-th* rule. $M(X_T,x_k)$ stands for a similarity measure, which can be defined as any similarity measure. The bottleneck problem of (8) is however caused by its requirement to revisit already seen samples. This issue can be tackled by formulating the recursive expression of (8). Instead of formulating the recursive definition of (8), the spatial firing strength of the data cloud suffices to be an alternative because it is derived from the idea of local density and is computed based on the local mean [1] which summarizes the characteristic of data streams. (8) is then written as follows:

$$P(X_t \in N_i) = \frac{\Lambda_i}{\sum_{i=1}^{R} \Lambda_i} \tag{9}$$

where $\Lambda_i$ is a type-reduced activation degree $\Lambda_i = (1-q)\overline{G}_{i,spatial} + q\underline{G}_{i,spatial}$. Once the neighbourhood probability is determined, its entropy is formulated as follows:

$$H(\mathrm{N}|X_t) = -\sum_{i=1}^{R} P(X_t \in N_i)\log P(X_t \in N_i) \quad (10)$$

The entropy of the neighbourhood probability measures the uncertainty induced by a training pattern. A sample with high uncertainty should be admitted for the model update, because it cannot be well-covered by an existing network structure. Learning such a sample is beneficial, because it minimises uncertainty. A sample is to be accepted for model updates, provided that the following condition is met:

$$H \geq thres \quad (11)$$

where $thres$ is an uncertainty threshold. This parameter is not fixed during the training process, rather it is dynamically adjusted to suit the learning context. This strategy is necessary to compensate for the potential increase of training samples to be accepted during the presence of concept drift. The threshold is set as $thres_{N+1} = thres_N(1 \pm inc)$, where it augments $thres_{N+1} = thres_N(1 + inc)$ when a sample is admitted for the training process, whereas it decreases $thres_{N+1} = thres_N(1 - inc)$ when a sample is ruled out for the training process. *inc* here is a step size, set at *inc=0.01*. This simply follows its default setting in [68].

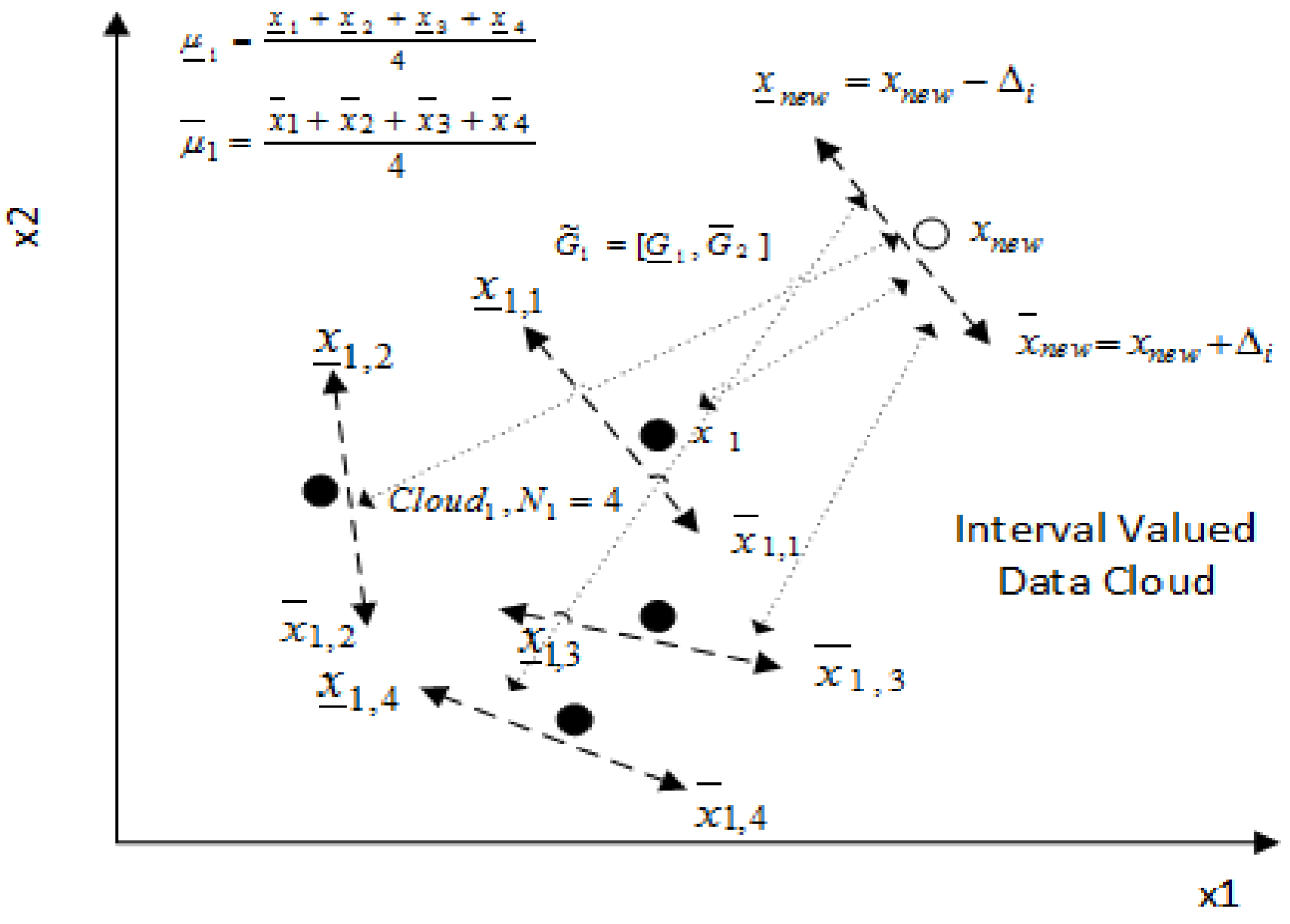

Fig. 2 Interval Valued Data Cloud

### 3.2 *Hidden Node Growing Strategy*

pRVFLN relies on the T2SCC method to grow interval-valued data clouds on demands. This notion is extended from the so-called SCC method [26], [65] to be well-suited with the type-2 hidden node working framework. The significance of the hidden nodes in pRVFLN is evaluated by checking its input and output coherence through analysis of its correlation to existing data clouds and the target concept. Let $\tilde{\mu}_i = [\underline{\mu}_i, \overline{\mu}_i] \in \Re^{1\times n}$ is a local mean of the *i-th* interval-valued data cloud (5), $X_t \in \Re^n$ is an input vector and $T_t \in \Re^m$ is a target vector, the input and output coherence are written as follows:

$$I_C(\tilde{\mu}_i, X_t) = (1-q)\zeta(\overline{\mu}_i, X_t) + q\zeta(\underline{\mu}_i, X_t) \quad (12)$$

$$O_C(\tilde{\mu}_i, X_t) = (\zeta(X_t, T_t) - \zeta(\tilde{\mu}_i, T_t)), \zeta(\tilde{\mu}_i, T_t) = (1-q)\zeta(\overline{\mu}_i, T_t) + q\zeta(\underline{\mu}_i, T_t) \quad (13)$$

where $\zeta()$ is the correlation measure. Both linear and non-linear correlation measures are applicable here. However, the non-linear correlation measure is rather hard to deploy in the online environment, because it is usually executed using the Discretization or Parzen Window method [49]. This often leads to an assumption of uniform data distribution as implemented in the differential entropy [28]. The Pearson correlation measure is a widely used correlation measure but it is insensitive to the scaling and translation of variables as well as being sensitive to rotation [35]. The maximal information compression index (MCI) is one of attempts to tackle these problems and is used in the T2SCC to perform the correlation measure $\zeta()$ [35]. It is defined as follows:

$$\zeta(X_1,X_2)=\frac{1}{2}(\mathrm{var}(X_1)+\mathrm{var}(X_2)-\sqrt{(\mathrm{var}(X_1)+\mathrm{var}(X_2))^2-4\,\mathrm{var}(X_1)\,\mathrm{var}(X_2)(1-\rho(X_1,X_2)^2)}) \quad (14)$$

$$\rho(X_1,X_2)=\frac{\mathrm{cov}(X_1,X_2)}{\sqrt{\mathrm{var}(X_1)\,\mathrm{var}(X_2)}} \quad (15)$$

where *(X1,X2)* are to be substituted by $(\bar{\mu}_i,X_t),(\underline{\mu}_i,X_t),(\bar{\mu}_i,T_t),(\underline{\mu}_i,T_t),(\bar{\mu}_i,X_t),(X_t,T_t)$. $\mathrm{var}(X),\mathrm{cov}(X_1,X_2),\rho(X_1,X_2)$ respectively stand for the variance of X, covariance of $X_1$ and $X_2$, and Pearson correlation index of $X_1$ and $X_2$. The local mean of the interval-valued data cloud is used to represent a data cloud because it captures the influence of all intervals of a data cloud. In essence, the MCI method indicates the amount of information compression when ignoring a newly observed sample. The principal component direction is referred to here, because it signifies the maximum information compression, resulting in maximum cost to be imposed when ignoring a datum. The MCI method features the following properties: 1) $0\le\zeta(X_1,Y_2)\le 0.5(\mathrm{var}(X_1)+\mathrm{var}(X_2))$, 2) a maximum correlation is given by $\zeta(X_1,X_2)=0$, 3) a symmetric property $\zeta(X_1,X_2)=\zeta(X_2,X_1)$, 4) the mean expression is discounted here. This makes it invariant against the translation of the dataset, and 5) it is also robust against rotation, which is verifiable from the perpendicular distance of a point to a line, and is unaffected by the rotation of the input features.

The input coherence explores the similarity between new data and existing data clouds directly, while the output coherence focusses on their dissimilarity indirectly through a target vector as a reference. The input and output coherence formulates a test that determines the degree of confidence to the current's hypothesis:

$$I_C(\tilde{\mu}_i,X_t)\le\alpha_1, O_C(\tilde{\mu}_i,X_t)\ge\alpha_2 \quad (16)$$

where $\alpha_1\in[0.001,0.01],\alpha_2\in[0.01,0.1]$ are predefined thresholds. If a hypothesis meets both conditions, a new training sample is assigned to a data cloud with the highest input coherence $i*$. Accordingly, the number of intervals $N_{i*}$, local mean and square length $\tilde{\mu}_{i*},\tilde{\Sigma}_{i*}$ are updated respectively with (5) and (6) as well as $N_{i*}=N_{i*}+1$. A new data cloud is introduced, provided that none of the existing hypotheses pass the test (16) – one of the conditions is violated. This situation reflects the fact that a new training pattern conveys significant novelty, which has to be incorporated to enrich the scope of the current hypotheses. Note that a larger $\alpha_1$ is specified, fewer data clouds are generated and vice versa, whereas a larger $\alpha_2$ is specified, larger data clouds are added and vice versa. The sensitivity of these two parameters are studied in the other section of this paper. Because a data cloud is a non-parametric model, no parameterization is committed for a new data cloud. The output node of a new data cloud is initialised:

$$W_{R+1} = W_{i*}, \Psi_{R+1} = \varpi I \tag{17}$$

where $\varpi = 10^5$ is a large positive constant. The output node is set as the data cloud with the highest input coherence because this data cloud is the closest one to the new data cloud. Furthermore, the setting of covariance matrix can produce a close approximation of the global minimum solution of batched learning, as proven mathematically in [31].

*3.3 Hidden Node Pruning and Recall Strategy*

pRVFLN incorporates a data cloud pruning scenario, termed the type-2 relative mutual information (T2RMI) method. This method was firstly developed in [21] for the type-1 fuzzy system and extended in [41] to adapt to the interval-valued working principle. This method is convenient to use in pRVFLN because it estimates mutual information between a data cloud and a target concept by analysing their correlation. Hence, the MCI method (14) - (15) can be applied to measure the correlation. Although this method has been well-established [21], [41], its effectiveness in handling data clouds and a recurrent structure as implemented in pRVFLN is to date an open question. Unlike both the RMI method [36] and T2RMI method [41] that apply the classic symmetrical uncertainty method, the T2RMI method is formulated using the MCI method as follows:

$$\zeta(\tilde{G}_{i,temp}, T_t) = q\zeta(\underline{G}_{i,temp}, T_t) + (1-q)\zeta(\overline{G}_{i,temp}, T_t) \tag{18}$$

where $\underline{G}_{i,temp}$ is a lower temporal activation function of the *i-th* rule. The temporal activation function is included in (17) rather than the spatial activation function in order for the inter-temporal dependency of the recurrent structure to be considered. The MCI method is chosen here, because it offers a good trade-off between accuracy and simplicity. It possesses significantly lower computational burden than the symmetrical uncertainty method even when implemented with the differential entropy [28] but is more robust than a widely used Pearson correlation index. A data cloud is deemed inconsequential and thus is able to be removed with negligible impact to accuracy, if (19) is met:

$$\zeta_i < mean(\zeta_i) - 2std(\zeta_i) \tag{19}$$

where $mean(\zeta_i), std(\zeta_i)$ are respectively the mean and standard deviation of the MCI during its lifespan. This criterion aims to capture an obsolete data cloud, which does not keep up with current data distribution due to possible concept drift, because it computes the downtrend of the MCI during its lifespan. It is worth mentioning that mutual information between hidden nodes and the target variable is a reliable indicator for changing data distributions, because it is in line with the definition of concept drift. Concept drift refers to a situation, where the posterior probability changes overtime $P(T_t|X_t) \neq P(T_{t-1}|X_{t-1})$.

The T2RMI method also functions as a rule recall mechanism, which is capable of coping with cyclic concept drift. Cyclic concept drifts frequently happen in the weather, customer preference, electricity power consumption problems, etc. where seasonal change comes into picture. This points to a situation where previous data distribution reappears again in the current training step. Once pruned by the T2RMI, a data cloud is not forgotten permanently and is inserted into a list of pruned data clouds $R^*=R^*+1$. In this case, its local mean, square length, population, an output node, and output covariance matrix $\tilde{\mu}_{R^*}, \tilde{\Sigma}_{R^*}, N_{R^*}, \beta_{R^*}, \Psi_{R^*}$ are retained in the memory. Such data clouds can be reactivated again in the future, whenever its validity is confirmed by an up-to-date data trend. It is worth noting that adding a completely new data cloud when observing previously learned concept violates the notion of an evolving learner and catastrophically erases learning history. A data cloud is recalled subject to the following condition:

$$\max_{i^*=1,..,R^*}(\zeta_{i^*}) > \max_{i=1,..,R}(\zeta_i) \tag{20}$$

This situation reveals that a previously pruned data cloud is more relevant than any existing ones. This condition pinpoints that a previously learned concept may reappear again. A previously pruned data cloud is then regenerated as follows:

$$\tilde{\mu}_{R+1} = \tilde{\mu}_{R^*}, \tilde{\Sigma}_{R+1} = \tilde{\Sigma}_{R^*}, N_{R+1} = N_{R^*}, \beta_{R+1} = \beta_{R^*}, \Psi_{R+1} = \Psi_{R^*} \tag{21}$$

Note that although previously pruned data clouds are stored in the memory, the data cloud pruning module still contributes in lowering the computational load, because all previously pruned data clouds are excluded from any training scenarios except (17). Unlike its predecessors [39], this rule recall scenario is not involved in the data cloud growing process (please refer to Algorithm 1) and plays a role as another data cloud generation mechanism. This mechanism is also developed from the T2RMI method, which can represent the change of posterior probability – concept drift - more accurately than the density concept.

### *3.4 Online Feature Selection Strategy*

Although feature selection and extraction problems have attracted considerable research attention; little effort has been paid toward online feature selection. Two common approaches to tackling this issue are through soft or hard dimensionality reduction techniques [39], [46]. Soft dimensionality reduction minimizes the effect of inconsequential features by assigning low weights but still retains a complete set of input attributes in the memory, whereas hard dimensionality reduction lowers the input dimension by cutting off spurious input features. Nonetheless, the hard dimensionality reduction method undermines stability, because an input feature cannot be retrieved once pruned [5]. To date, most of existing works always start the input selection process from a full set of input attributes and gradually reduces the number as more observation are come across. A prominent work, namely online feature selection (OFS), was developed in [3] and covers both partial and full input conditions. The appealing trait of OFS lies in its aptitude for flexible feature selection, where it enables the provision of different combinations of input attributes in each episode by activating or deactivating input features (1 or 0), which adapts to up-to-date data trends. Furthermore, this technique is also capable of handling partial input attributes which happens to be fruitful when the cost of feature extraction is too expensive. OFS was originally devised for the LR and is generalized here to fit the context of pRVFLN.

We start our discussion from a condition where a learner is provided with full input variables. Suppose that $B$ input attributes are to be selected in the training process and $B<n$, the simplest approach is to discard the input features with marginal accumulated output weights $\sum_{i=1}^{R}\sum_{j=1}^{2}\beta_{i,j}$ and maintain only $B$ input features with the largest output weights. Note that the second term $\sum_{j=1}^{2}$ is required because of the extended input vector $x_e \in \Re^{(2n+1)}$. The rule consequent informs a tendency of a rule which can be used as an alternate of gradient information which changes in each point [35]. Although it is straightforward to use, it cannot ensure the stability of the pruning process due to a lack of sensitivity analysis of the feature contribution. To correct this problem, a sparsity property of the L1 norm can be analyzed to examine whether the values of $n$ input features are concentrated in the L1 ball. This allows the distribution of the input values to be checked to determine whether they are concentrated in the largest elements and that pruning the smallest elements won't harm the model's accuracy. This concept is actualized by first inspecting the accuracy of pRVFLN. The input pruning process is carried out when the system error is large enough $T_t - y_t > \kappa$ or, in the realm of the classification problem, misclassification is made. Nevertheless, the system error is not only large in the case of underfitting, but also in the case of overfitting. We modify this condition by taking into account the evolution of system error $|\bar{e}_t + \sigma_t| > \kappa|\bar{e}_{t-1} + \sigma_{t-1}|$. The constant $\kappa$ is a predefined parameter and fixed at 1.1 for simplicity. The output nodes are updated using the gradient descent approach and then projected to the $L_2$ ball to guarantee a bounded norm. Algorithm 2 details the algorithmic development of pRVFLN.

**Algorithm 2.** GOFS using full input attributes

*Input* : α learning rate, χ regularization factor, $B$ the number of features to be retained

*Output*: selected input features $X_{t,selected} \in \Re^{1\times B}$

**For** *t=1,…., T*

Make a prediction $y_t$

**IF** $|\bar{e}_t + \sigma_t| > 1.1|\bar{e}_{t-1} + \sigma_{t-1}|$ // for regression or $\hat{o} = \max_{o=1,..,m}(y_o) \neq T_t$ // for classification

$\beta_i = \beta_i - \chi\alpha\beta_i - \alpha\chi\frac{\partial E}{\partial \beta_i}$, $\beta_i = \min(1, \frac{1/\sqrt{\chi}}{\|\beta_i\|_2})\beta_i$

Prune input attributes $X_t$ except those of $B$ largest $\sum_{i=1}^{R}\sum_{j=1}^{2}\beta_{i,j}$

**Else**

$\beta_i = \beta_i - \chi\alpha\beta_i$

**End IF**

**End FOR**

where $\alpha, \chi$ are respectively the learning rate and regularization factor. We assign $\alpha = 0.2, \chi = 0.01$ following the same setting as [3]. The optimization procedure relies on the standard mean square error (MSE) as the objective function, which leads to the following gradient term:

$$\frac{\partial E}{\partial \beta_i} = -(T_t - y_t)\left\{\sum_{i=1}^{R}(1-q)\overline{G}_{i,temporal} + \sum_{i=1}^{R} q\underline{G}_{i,temporal}\right\}$$

Furthermore, the system error has been theoretically proven to be bounded in [59] and the upper bound is also found. One can also notice that the GOFS enables different feature subsets to be elicited in each training observation *t*.

A relatively unexplored area of existing online feature selection is a situation where a limited number of features is accessible for the training process. To actualise this scenario, we assume that at most *B* input variables can be extracted during the training process. This strategy, however, cannot be done by simply acquiring any *B* input features, because this scenario results in the same subset of input features during the training process. This problem is addressed using the Bernaoulli distribution with confidence level $\varepsilon$ to sample *B* input attributes from *n* input attributes *B*<*n*. Algorithm 3 displays the feature selection procedure.

**Algorithm 3.** GOFS using partial input attributes

*Input* : α learning rate, χ regularization factor, *B* the number of features to be retained, $\varepsilon$ confidence level

*Output*: selected input features $X_{t,selected} \in \Re^{1\times B}$

**For** *t=1,…., T*

Sample $\gamma$ from Bernaoulli distribution with confidence level $\varepsilon$

**IF** $\gamma_t = 1$

Randomly select *B* out of *n* input attributes $\tilde{X}_t \in \Re^{1\times B}$

**End IF**

Make a prediction $y_t$

**IF** $|\overline{e}_t + \sigma_t| > 1.1|\overline{e}_{t-1} + \sigma_{t-1}|$ // for regression or $\hat{o} = \max_{o=1,..,m}(y_o) \neq T_t$ // for classification

$\hat{X}_t = \tilde{X}_t / (B/n\varepsilon) + (1-\varepsilon)$

$\beta_i = \beta_i - \chi\alpha\beta_i - \alpha\chi \frac{\partial E(\hat{X}_t)}{\partial \beta_i}$, $\beta_i = \min(1, \frac{1/\sqrt{\chi}}{\|\beta_i\|_2})\beta_i$

Prune input attributes $X_t$ except those of $B$ largest $\sum_{i=1}^{R}\sum_{j=1}^{2}\beta_{i,j}$

**Else**

$\beta_{i,t} = \beta_{i,t-1}$

**End IF**

**End FOR**

As with Algorithm 2, the convergence of this scenario has been theoretically proven and the upper bound is derived in [3]. One must bear in mind that pruning process in Algorithm 1, 2 are carried out by assigning crisp weights (0 or 1), which fully reflects the importance of input features.

*3.5 Random Learning Strategy*

pRVFLN adopts the random parameter learning scenario of the RVFLN leaving only the output nodes $W$ to be analytically tuned with an online learning scenario, whereas others, namely $A_t, q, \lambda, \Delta$, can be randomly generated without any tuning process. To begin the discussion, we recall the output expression of pRVFLN as follows:

$$y_o = \sum_{i=1}^{R} \beta_i \tilde{G}_{i,temporal}(X_t; A_t, q, \lambda, \Delta) \quad (22)$$

Note that the pRVFLN possesses the enhancement layer expressed as a nonlinear mapping using the Chebyshev function $\beta_i = x_e w_i$ as with the original RVFLN [48]. Referring to the RVFLN theory, the hidden node $\tilde{G}_{i,spatial}$ and its derivative must be integrable:

$$\int_R G^2(x)dx < \infty, or \int_R [G'(x)]^2 dx < \infty \quad (23)$$

Furthermore, a large number of hidden nodes $R$ is usually selected to ensure adequate coverage of hidden nodes to a data space because they are chosen at random [41]. Nevertheless, this condition can be relaxed in the pRVFLN, because the data cloud growing mechanism, namely the T2SCC method, partitions the input region in respect to real data distributions. The concept of data cloud-based neurons features the concept of local density, which adapts to any variation of data streams. Furthermore, this concept is parameter-free and thus does not require any parameterization, which calls for a high-level approximation or complicated optimization procedure. Other parameters, namely $A_t, q, \lambda, \Delta$, are randomly chosen, and their region of randomisation should be carefully selected. Referring to Ingelnik and Pao [20], the random parameters are sampled randomly from the following.

$$\begin{cases} w \in [0, \tau Z] \times [-\tau Z, \tau Z]^n \\ \kappa \in [0,1]^n \\ \mu \in [-2nZ, 2nZ] \\ b = -w\kappa - \mu \end{cases} \tag{24}$$

where $\tau, Z, \mu$ are probability measures. Nevertheless, this strategy is impossible to implement in the online situations, because it often entails a rigorous trial-error process to determine these parameters. Most RVFL works simply follow Schmidt et al.'s strategy [53], setting the region of random parameters in the range of [-1,1].

Assuming that a complete dataset $\Xi = [X, T] \in \Re^{N \times (n+m)}$ is observable, a closed-form solution of (22) can be defined to determine the output weights $w$. Although the original RVFLN adjusts the output weight with the conjugate gradient (CG) method, the closed-form solution can be still utilised with ease [48]. The mere obstacle for the use of pseudo-inversion in the original work was the issue of limited computational resources in 90's. Note that the regularization technique needs to be undertaken if the hidden node matrix is ill-conditioned. Although it is easy to use and ensures a globally optimum solution, this parameter learning scenario however imposes revisiting preceding training patterns which is intractable for online learning scenarios. pRVFLN employs the FWGRLS method [35], [39] to adjust the output weight which is underpinned by a generalized weight decay function. Because the FWGRLS approach has been detailed in [39], it is not recounted here.

### *3.6 Robustness of RVFLN*

The network parameters are usually sampled uniformly within a range of [-1,1] in the literature. A new finding of Li and Wang in [29] exhibits that the process of randomly generating network parameters with a fixed scope [-α, α] does not ensure a theoretically feasible network or often the hidden node matrix is not full rank. Surprisingly, the hidden node matrix was not invertible in all their cases when randomly sampling network parameters in the range of [-1,1] and far better numerical results were achieved by choosing the scope [-200,200]. This trend was consistent with different numbers of hidden nodes. How to properly select scopes of random parameters and its corresponding distribution still require in-depth investigation [53]. In practice, a pre-training process is normally required to arrive at decent local mapping. Note that the range of random parameters by Ingelnik and Pao [20] is still at the theoretical level and does not touch implementation issue. We study different random regions in Section IV. C to see how pRVFLN behaves under variations of scope of random parameters.

## IV. Numerical Examples

This section presents numerical validations of our proposed algorithm using case studies and comparisons with prominent algorithms in the literature. Two numerical examples, namely modelling of Nox emissions from a car engine, tool condition monitoring in the ball-nose end milling process, are presented in this section, and the other three numerical examples are placed in the supplemental

document[1)] to keep the paper compact. Our numerical studies were carried out under two scenarios: the time-series scenario and the cross-validation scenario. The time-series procedure orderly executes data streams according to their arrival time and partitions data streams into two parts, namely training and testing. In the time-series mode, pRVFLN was compared against 10 state-of-the-art evolving algorithms: eT2Class [40], BARTFIS [37], PANFIS [42], GENEFIS [43], eTS+ [5], eTS [6], simp_eTS [7], DFNN [61], GDFNN [62], FAOSPFNN [60], ANFIS [25]. The CV scenarios were taken place in our experiment in order to follow the commonly adopted simulation environment of other RVFLNs in the literature. The pRVFLN was benchmarked against 8 prominent RVFLNs in the literature: decorelated neural network ensemble (DNNE) [8], centralized RVFL (A) [51], RLS-consensus-RVFL [51] (B), RLS-local-RVFL (C) [51], LMS-concensus-RVFL (D) [51]. The MATLAB codes of these algorithms are available online [2,3)]. Comparisons were performed against five evaluation criteria: accuracy, data clouds, input attribute, runtime, network parameters. The scope of random parameters followed Schmidt's suggestion [53] where the scope of random parameters was [-1,1] but we insert analysis of robustness in part C which provides additional results with different random regions. The effect of individual learning component to the end results and the influence of user-defined predefined thresholds are analysed in the supplemental document[1)]. To allow a fair comparison, all consolidated algorithms were executed in the same computational resources under the MATLAB's environment.

Table 1. Prediction of Nox emissions Using Time-Series Mode

| Model | RMSE | Node | Input | Runtime | Network | Samples |
|---|---|---|---|---|---|---|
| pRVFLN (P) | **0.04** | **2** | 5 | **0.24** | 22 | **510** |
| pRVFLN (F) | 0.05 | **2** | 5 | 0.56 | 22 | 515 |
| eT2Class | 0.045 | 2 | 170 | 17.98 | 117304 | 667 |
| PANFIS | 0.052 | 5 | 170 | 3.37 | 146205 | 667 |
| GENEFIS | 0.048 | 2 | **2** | 0.41 | **18** | 667 |
| Simp_eTS | 0.14 | 5 | 170 | 5.5 | 1876 | 667 |
| BARTFIS | 0.11 | 4 | 4 | 2.55 | 52 | 667 |
| DFNN | 0.18 | 548 | 170 | 4332.9 | 280198+NS | 667 |
| GDFNN | 0.48 | 215 | 170 | 2144.1 | 109865 | 667 |
| eTS | 0.38 | 27 | 170 | 1098.4 | 13797 | 667 |
| FAOS-PFNN | 0.06 | 6 | 170 | 14.8 | 2216+NS | 667 |
| ANFIS | 0.15 | 2 | 170 | 100.41 | 17178 | 667 |
| eTS+ | 0.14 | 4 | 4 | a) | 28 | 667 |

a): the result was obtained under different computer platform

Table 2. Prediction of Nox emissions Using CV Mode

| Model | NRMSE | Node | Input | Runtime | Network | Samples |
|---|---|---|---|---|---|---|
| pRVFLN(P) | **0.12±0.04** | **2** | **5** | 1.1 | **22** | 699 |
| pRVFLN(F) | 0.12±0.03 | **2** | **5** | 5.98 | **22** | **630.7** |
| DNNE | 0.14±0 | 50 | 170 | 0.81 | 43600+NS | 744 |
| A-online | 0.52±0.02 | 100 | 170 | 0.03 | 87200 | 744 |
| B-online | 0.57±0.01 | 100 | 170 | 0.41 | 87200 | 744 |
| C-online | 0.93±0.02 | 100 | 170 | 0.21 | 87200 | 744 |
| D-online | 1.7±0.04 | 100 | 170 | 0.09 | 87200 | 744 |
| A-batch | 0.59±0.05 | 100 | 170 | 0..003 | 87200+NS | 744 |
| B-batch | 0.62±0.05 | 100 | 170 | **0.001** | 87200+NS | 744 |
| C-batch | 0.83±0.09 | 100 | 170 | **0.001** | 87200+NS | 744 |

*A. Modeling of Nox Emissions from a Car Engine*: this section demonstrates the efficacy of the pRVFLN in modeling Nox emissions from a car engine [32]. This real-world problem is relevant to validate the learning performance, not only because it features noisy and uncertain characteristics as the nature of a car engine, but also it characterizes high dimensionality, containing 170 input attributes. That is, 17 physical variables were captured in 10 consecutive measurements. Furthermore, different engine parameters were applied to induce changing system dynamic to simulate real driving conditions. In the time-series procedure, 826 data points were streamed to consolidated algorithms, where 667 samples were

[1)] https://www.dropbox.com/s/lytpt4huqyoqa6p/supplemental_document.docx?dl=0
[2)] http://homepage.cs.latrobe.edu.au/dwang/html/DNNEweb/index.html
[3)] http://ispac.ing.uniroma1.it/scardapane/software/lynx/.

set as training samples, and the remainder were fed for testing purposes. In the CV procedure, the experiment was run under the 10-fold CV scheme, and each fold was repeated five times similar to the scenario adopted in [22]. This strategy checked the consistency of the RVFLN's learning performance because it adopts the random learning scenario and avoids data order dependency. Table 1 and 2 exhibit consolidated numerical results of benchmarked algorithms.

Table 3. Tool Wear Prediction Using Time Series Mode

| Model | RMSE | Node | Input | Runtime | Network | Samples |
|---|---|---|---|---|---|---|
| pRVFLN (P) | **0.14** | **2** | **8** | 0.34 | **34** | 157 |
| pRVFLN (F) | 0.19 | **2** | **8** | **0.15** | **34** | **136** |
| eT2Class | 0.16 | 4 | 12 | 1.1 | 1260 | 320 |
| Simp_eTS | 0.22 | 17 | 12 | 1.29 | 437 | 320 |
| eTS | 0.15 | 7 | 12 | 0.56 | 187 | 320 |
| BARTFIS | 0.16 | 6 | 12 | 0.43 | 222 | 320 |
| PANFIS | 0.15 | 3 | 12 | 0.77 | 507 | 320 |
| GENEFIS | 0.13 | 42 | 12 | 0.88 | 507 | 320 |
| DFNN | 0.27 | 42 | 12 | 2.41 | 1092+NS | 320 |
| GDFNN | 0.26 | 7 | 12 | 2.54 | 259+ NS | 320 |
| FAOS-PFNN | 0.38 | 7 | 12 | 3.76 | 1022+NS | 320 |
| ANFIS | 0.16 | 8 | 12 | 0.52 | 296+ NS | 320 |

Table 4. Tool wear prediction using CV Mode

| Model | NRMSE | Node | Input | Runtime | Network | Samples |
|---|---|---|---|---|---|---|
| pRVFLN(P) | 0.16±0.01 | **1.3±0.2** | **8** | 0.2 | **22.1** | **245.8** |
| pRVFLN(F) | 0.17±0.06 | 1.92±0.1 | **8** | 0.3 | 32.6 | 432.1 |
| DNNE | **0.11±0** | 50 | 12 | 0.79 | 3310+NS | 571.5 |
| A-online | 0.16±0.01 | 100 | 12 | 0.02 | 1400 | 571.5 |
| B-online | 0.21±0.01 | 100 | 12 | 0.05 | 1400 | 571.5 |
| C-online | 1.39±0.03 | 100 | 12 | 0.13 | 1400 | 571.5 |
| D-online | 0.25±0.01 | 100 | 12 | 0.09 | 1400 | 571.5 |
| A-batch | 0.19±0.04 | 100 | 12 | **0..002** | 1400+NS | 571.5 |
| B-batch | 0.21±0.03 | 100 | 12 | 0.009 | 1400+NS | 571.5 |
| C-batch | 0.23±0.05 | 100 | 12 | 0.007 | 1400+NS | 571.5 |

It is evident that pRVFLN outperformed its counterparts in all the evaluation criteria except GENEFIS for the number of input attributes and network parameters. It is worth mentioning however that in the other three criteria: predictive accuracy, execution time, and number of training samples, the GENEFIS was more inferior than ours. pRVFLN is equipped with the online active learning strategy, which discarded superfluous samples. This learning module had a significant effect of predictive accuracy. Furthermore, pRVFLN has the GOFS method, which was capable of coping with the curse of dimensionality. Note that the unique feature of the GOFS method is that it allows different feature subsets to be picked up in every training episode. This case avoids the catastrophic forgetting of obsolete input attributes, temporarily inactive due to changing data distributions. The GOFS can handle partial input attributes during the training process and resulted in the same level of accuracy as that of the full input attributes. The use of full input attributes slowed down the execution time because it needed to deal with 170 input variables first, before reducing the input dimension. In this case study, we selected five input attributes to be kept for the training process. Our experiment showed that the number of selected input attributes is not problem-dependent and able to be set as any desirable number in most cases. There was no significant performance difference when using either the full input mode or partial input mode. On the other hand, consistent numerical results were achieved by pRVFLN, although the pRVFLN is built on the random vector functional link algorithm, as observed in the CV experimental scenario. In addition, pRVFLN produced the most encouraging performance in almost all evaluation criteria but computational speed because the other RVFLNs, (A)-(D), implement less comprehensive training procedure than pRVFLN by only fine-tuning output weights without any structural learning and feature selection mechanisms.

*B. Tool Condition Monitoring of High-Speed Machining Process*: this section presents a real-world problem, taken from a complex manufacturing problem (Courtesy of Dr. Li Xiang, Singapore) [44]. The objective of this case study is to perform predictive analytics of the tool wear in the ball-nose end milling process frequently found in the metal removal process of the aerospace industry. In total, 12 time-domain features were extracted from the force signal and 630 samples were collected during the experiment. Concept drift in this case study was resulted from changing surface integrity, tool wear degradation as well as varying machining configurations. For the time-series experimental procedure, the consolidated algorithms were trained using data from cutter A, while the testing phase exploited data from cutter B. For the CV experimental procedure, the 10-fold CV process was undertaken where each fold was undertaken five times to arrive at consistent finding. Tables 3 and 4 report average numerical results across all folds. Fig. 3 depicts how many times input attributes are selected during one fold of CV process.

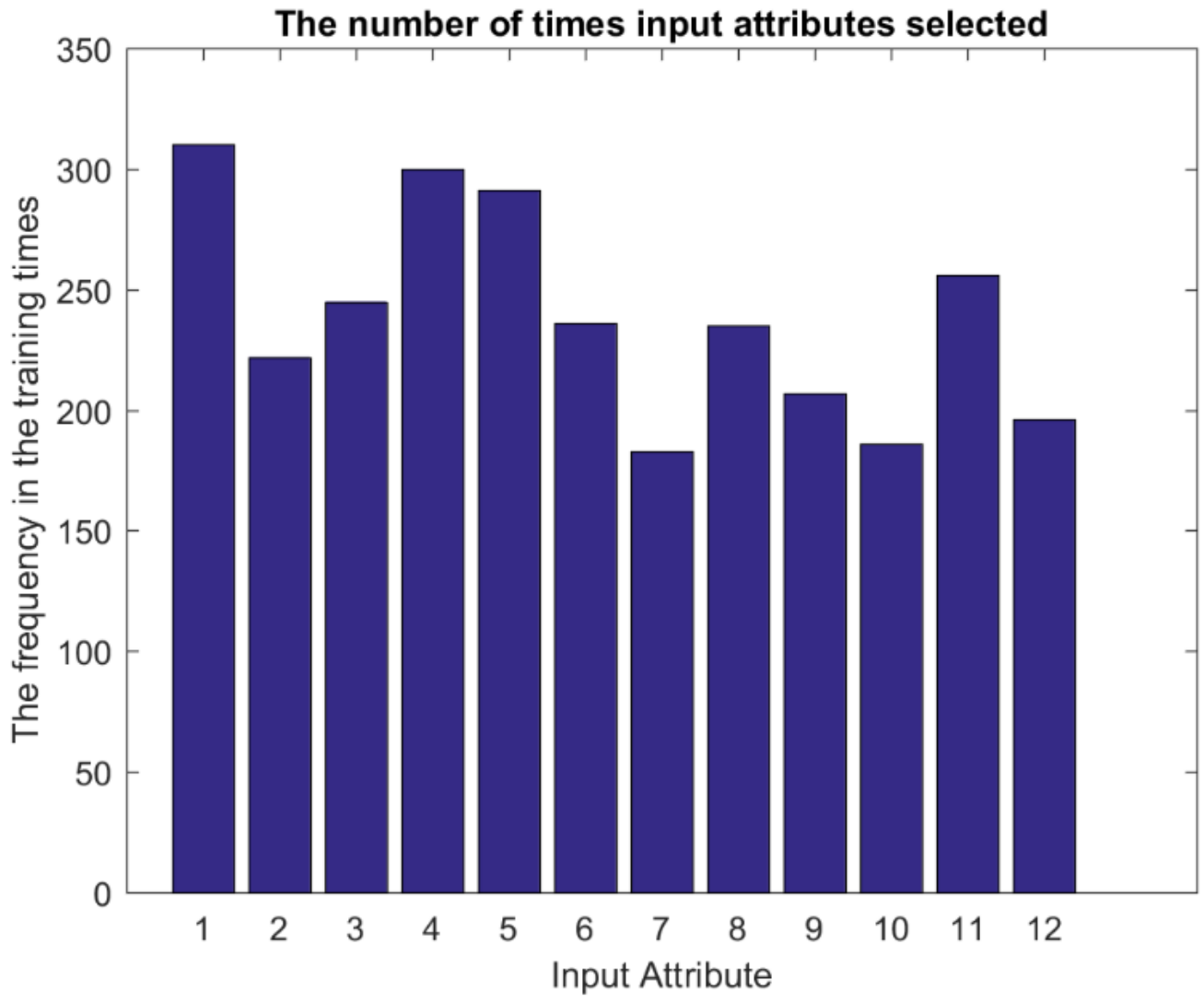

Fig. 3 the frequency of input features

It is observed from Table 3 and 4 that pRVFLN evolved the lowest structural complexities while retaining a very high level of accuracy. It is worth noting that although the DNNE exceeded pRVFLN in accuracy, it imposed considerable complexity because it is an offline algorithm revisiting previously seen data samples and adopts an ensemble learning paradigm. The efficacy of the online sample selection strategy was seen, as it led to significant reduction of training samples to be learned during the experiment. Using partial input information led to subtle differences to those with the full input information. It is seen in Fig. 3 that the GOFS selected different feature subsets in every training episode and it reflected the importance of input variables to the training process. Additional numerical examples, sensitivity analysis of predefined thresholds and analysis of learning modules can be found in [1)].

*C. Analysis of Robustness*: this section aims to numerically validate our claim in section III. E that a range [-1,1] does not always assure a reliable model [29],[53]. Additional numerical results with different intervals of random parameters are presented. Four intervals, namely [0,0.1],[0,0.5] [0,0.8] and [0,5] were tried out for two case studies in section IV.A and IV.B. Our experiments were undertaken in the 10-folds CV procedure where each fold was repeated five times to prevent the effect of random sampling. Our experiment made use of the feature selection scenario of a full number of input attributes. Table 5 displays numerical results.

Table 5. Analysis of Robustness

| Scope | Criteria | Tool Wear | Nox emission |
|---|---|---|---|
| [0,0.1] | RMSE | **0.13±0.008** | 1.9±0 |
| | Node | **1.8±0.25** | **1** |
| | Input | **8** | **5** |
| | Runtime | 0.2±0.1 | **0.1±0.02** |
| | Network | **30.9** | **11** |
| | Samples | 503.1 | **1** |
| [0,0.5] | RMSE | 0.14±0.02 | **0.1±0.01** |
| | Node | 1.92±0.2 | 1.98±0.14 |
| | Input | **8** | **5** |
| | Runtime | 0.18±0.008 | 5.7±0.3 |
| | Network | 32.6 | 21.8 |
| | Samples | 571.5 | 743.4 |
| [0,0.8] | RMSE | 0.47±0.42 | 0.18±0.3 |
| | Node | 1.4±0.05 | 1.96±0.19 |
| | Input | **8** | **5** |
| | Runtime | 0.19±0.13 | 5.56±0.96 |
| | Network | 23.8 | 21.6 |
| | Samples | **385.1** | 711.24 |
| [0,5] | RMSE | Unstable | Unstable |
| | Node | | |
| | Input | | |
| | Runtime | | |
| | Network | | |
| | Samples | | |

For the tool wear case study, the best-performing model was generated by the range [0,0.1]. The higher the range of the model the more inferior the model was. It went up to the point where a model was no longer stable under the range [0,5]. On the other side, the range [0,0.5] induced the best-performing model with the highest accuracy while evolving comparable network complexity for the Nox emission case study. The higher the scope led to the deterioration of numerical results. Moreover, the range [0,0.1] did not deliver a better accuracy than the range [0,0.5] either since this range did not generate diverse enough random values. These numerical results are interpreted from the nature of pRVFLN – a clustering-based algorithm. The success of pRVFLN is mainly determined from the compatibility of zone of influence of hidden nodes to a real data distribution, and its performance worsens when the scope is remote from the true data distribution. This finding is complementary to Li and Wang [29] where it relies on a sigmoid-based RVFL network, and the scope of random parameters can be outside the applicable operating intervals. Its predictive performance is set by its approximation capability in the output space.

# V. Conclusions

A novel random vector functional link network, namely parsimonious random vector functional link network (pRVFLN), is proposed. pRVFLN aims to provide a concrete solution to the issue of data stream by putting into perspective a synergy between adaptive and evolving characteristics and fast and easy-to-use characteristics of RVFLN. pRVFLN is a fully evolving algorithm where its hidden nodes can be automatically added, pruned and recalled dynamically while all network parameters except the output weights are randomly generated with the absence of any tuning mechanism. pRVFLN is fitted by the online feature selection mechanism and the online active learning scenario which further strengthens its aptitude in processing data streams. Unlike conventional RVFLNs, the concept of interval-valued data clouds is introduced. This concept simplifies the working principle of pRVFLN because it neither requires any parameterization per scalar variables nor follows pre-specified cluster shape. It features an interval-valued spatiotemporal firing strength, which provides the degree of tolerance for uncertainty. Rigorous case studies were carried out to numerically validate the efficacy of pRVFLN where pRVFLN delivered the most encouraging performance. The ensemble version of pRVFLN will be the subject of our future investigation which aims to further improve the predictive performance of pRVFLN.

## VI. ACKNOWLEDGEMENT

The third author acknowledges the support of the Austrian COMET-K2 program of the Linz Center of Mechatronics (LCM), funded by the Austrian federal government and the federal state of Upper Austria. We thank Dr. D. Wang for his suggestion pertaining to robustness issue of RVFLN.